%% file: main.tex
\documentclass[10pt,conference]{IEEEtran}
\IEEEoverridecommandlockouts

\usepackage{cite}
\usepackage{amsmath,amssymb,amsfonts}
\usepackage{algorithmic}
\usepackage{graphicx}
\usepackage{textcomp}
\usepackage[table,dvipsnames]{xcolor}

\usepackage{multirow}

\usepackage{float}

\ifCLASSOPTIONcompsoc
\usepackage[caption=false,font=normalsize,labelfont=sf,textfont=sf]{subfig}
\else
\usepackage[caption=false,font=footnotesize]{subfig}
\fi

\usepackage{tikz}
\usepackage{pgfplots}
\usetikzlibrary{spy, arrows, patterns}
\pgfdeclarelayer{bg}    
\pgfsetlayers{bg, main}  
\definecolor{redsynthesis}{HTML}{E76F51}
\definecolor{greenarm}{HTML}{2CA395}
\definecolor{purpleupsampling}{HTML}{6600CC}
\definecolor{bluelatent}{HTML}{829CBC}
\definecolor{orangenonoverfitted}{HTML}{FF9933}
\definecolor{orangecharted}{HTML}{FF7900}
\definecolor{graycharted}{HTML}{8F8F8F}
\definecolor{bluecharted}{HTML}{4BB4E6}
\definecolor{greencharted}{HTML}{50BE87}
\definecolor{pinkcharted}{HTML}{FFB4E6}
\definecolor{purplecharted}{HTML}{A885D8}
\definecolor{yellowcharted}{HTML}{FFD200}

\def\BibTeX{{\rm B\kern-.05em{\sc i\kern-.025em b}\kern-.08em
    T\kern-.1667em\lower.7ex\hbox{E}\kern-.125emX}}

\newcommand\imWidth{1.6in}

\begin{document}

\title{BOGausS: Better Optimized Gaussian Splatting}

\author{
    \IEEEauthorblockN{ 
        Stéphane Pateux\textsuperscript{1}, 
        Matthieu Gendrin\textsuperscript{1,2},
        Luce Morin\textsuperscript{2},
        Théo Ladune\textsuperscript{1},
        Xiaoran Jiang\textsuperscript{2}}
    \IEEEauthorblockA{
        \textsuperscript{1}\textit{Orange Innovation, Cesson Sévigné, France} \\        
        \textsuperscript{2}\textit{Univ Rennes, INSA Rennes, CNRS, IETR-UMR 6164, F-35000 Rennes, France} \\        
        name.surname@orange.com, name.surname@insa-rennes.fr}
}

\maketitle

\begin{abstract}
3D Gaussian Splatting (3DGS) proposes an efficient solution for novel view synthesis. Its framework 
provides fast and high-fidelity rendering. Although less complex than other solutions such as Neural
Radiance Fields (NeRF), there are still some challenges building smaller models without sacrificing 
quality. 
In this study, we perform a careful analysis of 3DGS training process and propose a
new optimization methodology.
Our Better Optimized Gaussian Splatting (BOGausS) solution is able to generate models 
up to ten times lighter than the original 3DGS with no quality degradation, 
thus significantly boosting the performance of Gaussian Splatting compared to the state of the art.
\end{abstract}


\input{1_intro}

\input{2_related}

\input{3_preliminary}

\input{4_methods}

\input{5_results}

\input{6_conclusion}

\bibliographystyle{IEEEtran}
\bibliography{main}

\end{document}

%% file: 1_intro.tex
\section{Introduction}
\label{sec:intro}

3D scene reconstruction has had significant improvements since the rise of Neural Radiance 
Fields (NeRF)~\cite{mipnerf21} and 3D Gaussian splatting (3DGS)~\cite{kerbl3Dgaussians},
opening the opportunity to many free view-point applications. 3DGS has gained high
popularity thanks to its high quality image rendering, reasonable training time and real 
time rendering. Among the various challenges associated to 3DGS is the optimization step 
of parameters estimation of the 3D Gaussians composing a scene. Several 
studies~\cite{MiniSplatting24,Taming24,MCMC24,Eagles24} propose to either improve 
3DGS rendering quality or reduce the number of Gaussians.

In this study, we propose to revisit the optimization process of 3DGS in order to jointly improve 
on model quality and compaction. 
First we propose a model to derive attainable precision reconstruction of a 3D scene. 
This precision model is then exploited to provide precision-aware updates during Stochastic Gradient 
Descent (SGD). 
We also propose a new unbiased optimizer based on Sparse Adam to improve SGD when dealing with 
sparsely-viewed and split Gaussians.
Targeting compactness of the model, we propose an improved densification mechanism with an analogy to 
Rate Distortion Optimization~\cite{RDO98}. 
Our main contributions are as follows: 
\begin{itemize}
  \item {\it Confidence estimation of 3D reconstruction}. Through a parallel with Variational Bayesian Inference, we propose to define
  confidence metrics on geometry reconstruction. From these findings a minimal size for 
  each Gaussian is defined depending on its localization. 
  \item {\it Unbiased Sparse Adam Optimizer}. We propose a revised Optimizer that ensures unbiased smooth 
  Stochastic Gradient Descent. This optimizer also exploits reconstruction confidence intervals to automatically
  adapt to scene content and camera setup.
  \item {\it Density preserving Gaussian splitting}. We propose a new splitting mechanism  
  that allows smooth loss transition. 
  \item {\it Distortion-based densification of Gaussians}. We propose densification and pruning strategy 
  inspired from rate-distortion optimization with new prioritization metrics.
\end{itemize}
Our proposed solution achieves 0.5-1 dB PSNR 
  increase over existing methods. Extensive experimental results and ablation study demonstrate the benefits
  of our proposed BOGausS method on several benchmarks and datasets.

\input{figure_rendered}

%% file: figure_rendered.tex
\begin{figure}[t] 
    \centering
    \subfloat[MCMC 0.51M - 22.51dB]{
        \begin{tikzpicture}
            \node[anchor=south west,inner sep=0] (image) at (0,0) {\includegraphics[width=\imWidth]{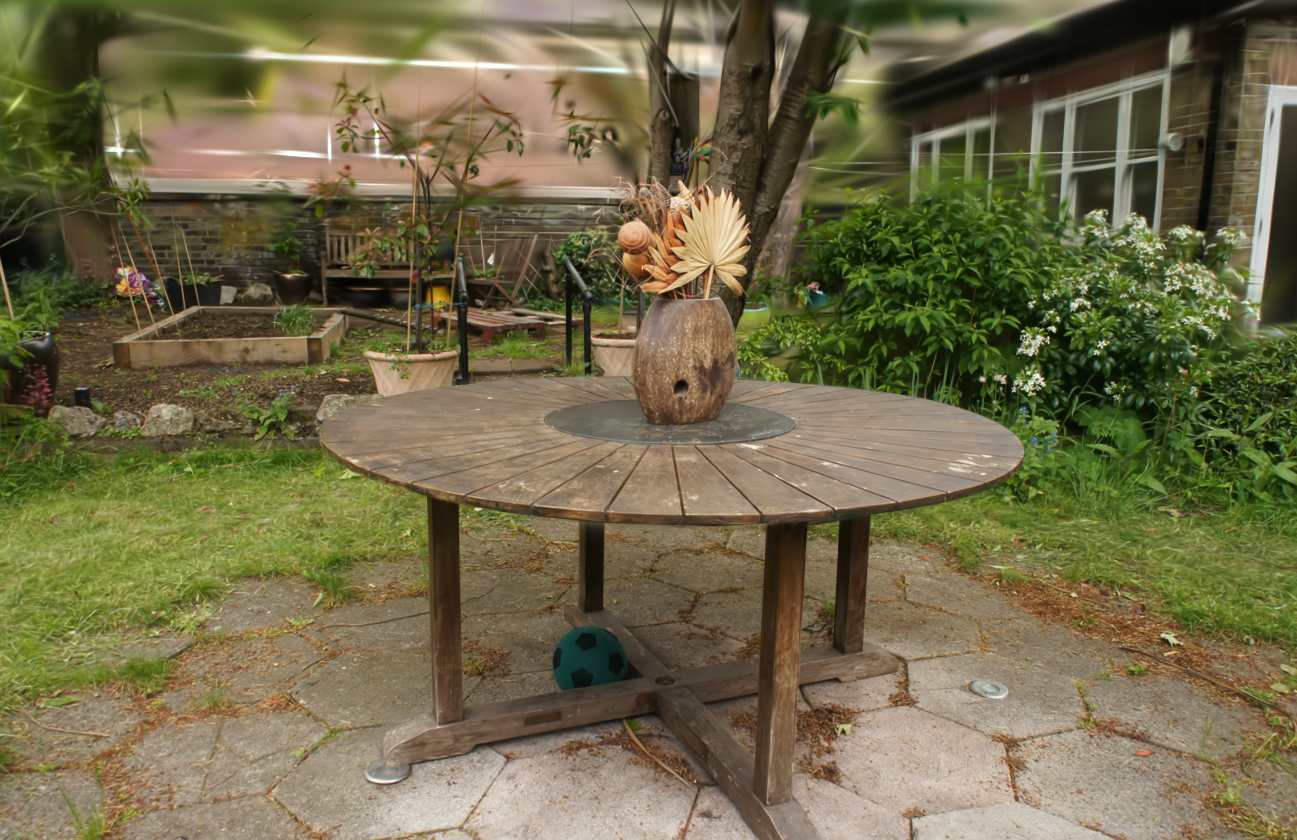}};
            \begin{scope}[x={(image.south east)},y={(image.north west)}]
                \draw[red,ultra thick,rounded corners] (0.01,0.70) rectangle (0.78,0.99);
            \end{scope}
        \end{tikzpicture}
        \label{fig:mcmc_low}
    }
    \hfil
    \subfloat[MCMC 4.06M - 23.38dB]{
        \begin{tikzpicture}
            \node[anchor=south west,inner sep=0] (image) at (0,0) {\includegraphics[width=\imWidth]{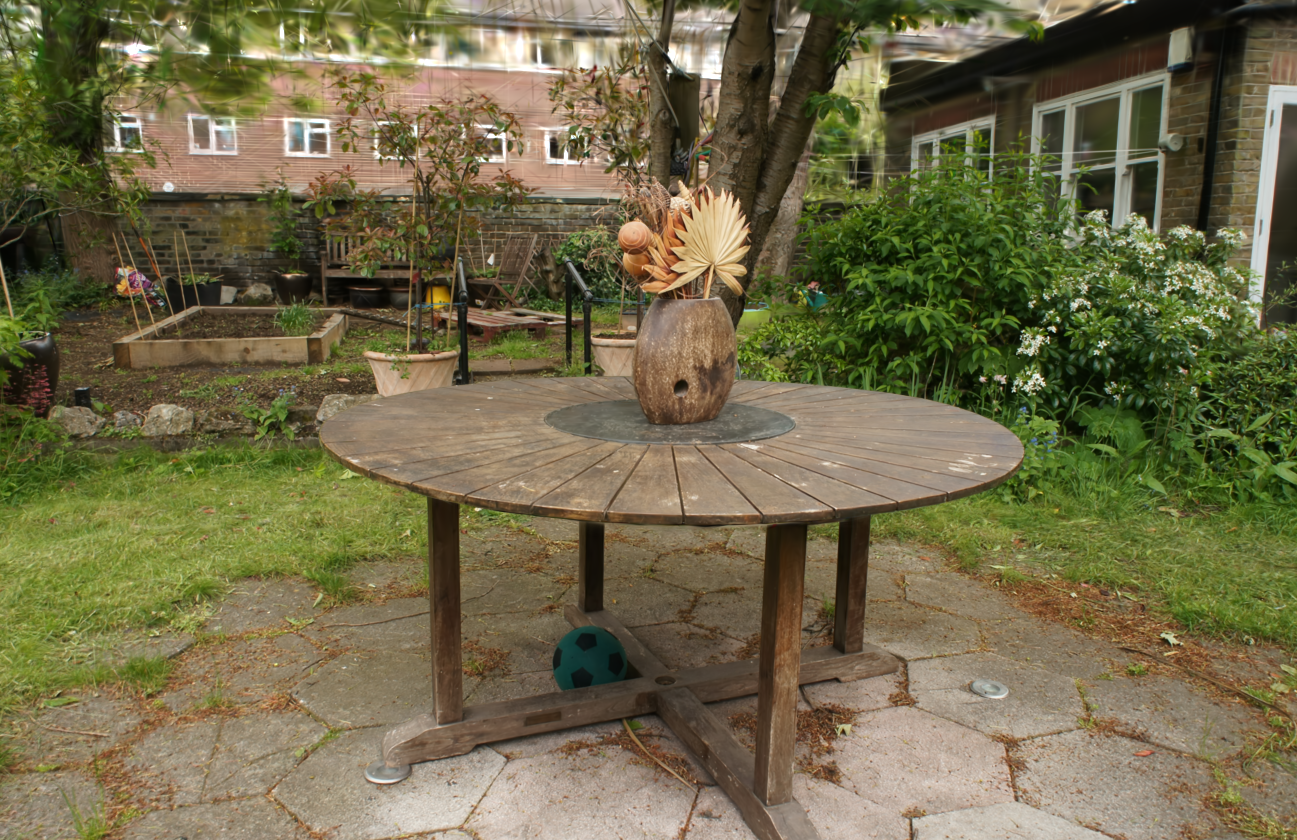}};
            \begin{scope}[x={(image.south east)},y={(image.north west)}]
                \draw[red,ultra thick,rounded corners] (0.01,0.70) rectangle (0.78,0.99);
            \end{scope}
        \end{tikzpicture}
        \label{fig:mcmc_high}
    }
    \hfil
    \subfloat[BOGausS 0.51M - 24.05dB]{
        \begin{tikzpicture}
            \node[anchor=south west,inner sep=0] (image) at (0,0) {\includegraphics[width=\imWidth]{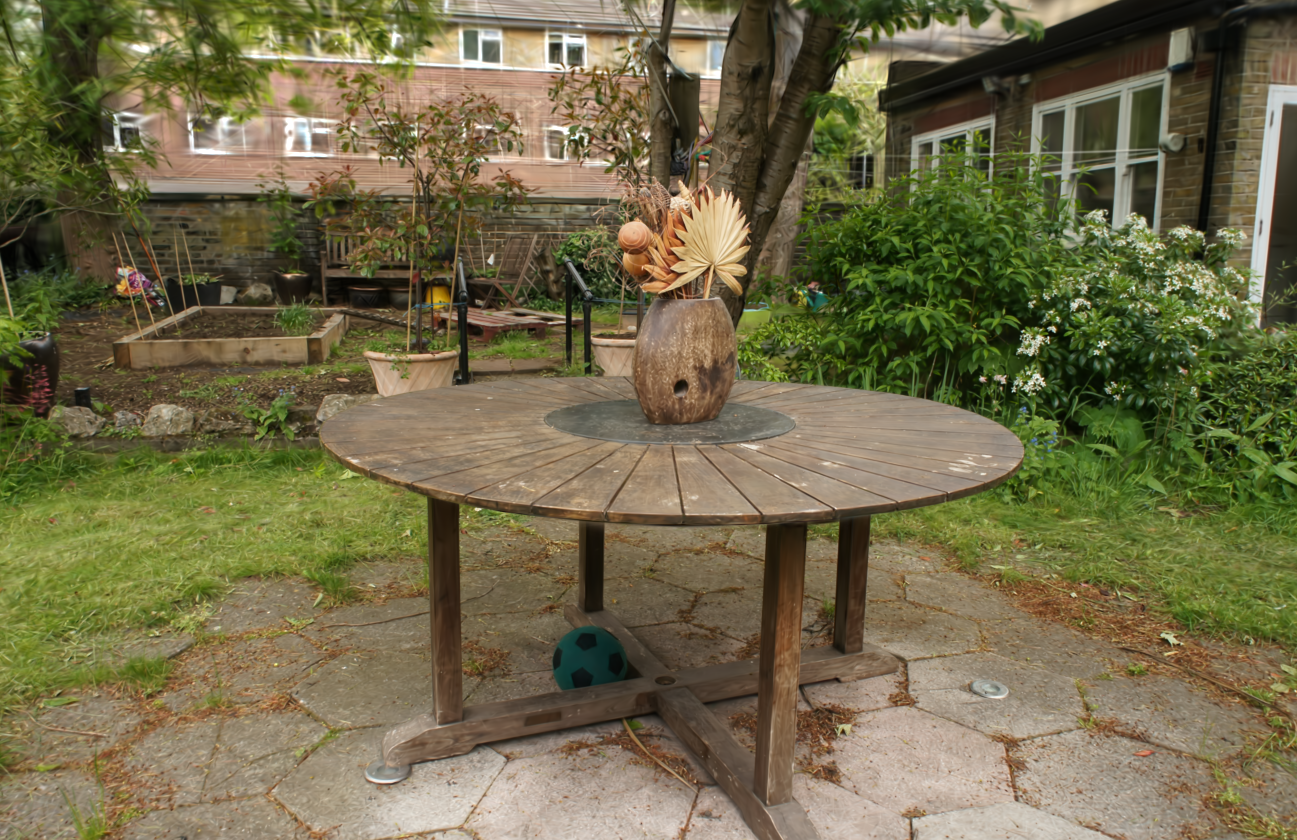}};
            \begin{scope}[x={(image.south east)},y={(image.north west)}]
                \draw[red,ultra thick,rounded corners] (0.01,0.70) rectangle (0.78,0.99);
            \end{scope}
        \end{tikzpicture}
        \label{fig:ours_low}
    }
    \hfil
    \subfloat[BOGausS 4.06M - 24.87dB]{
        \begin{tikzpicture}
            \node[anchor=south west,inner sep=0] (image) at (0,0) {\includegraphics[width=\imWidth]{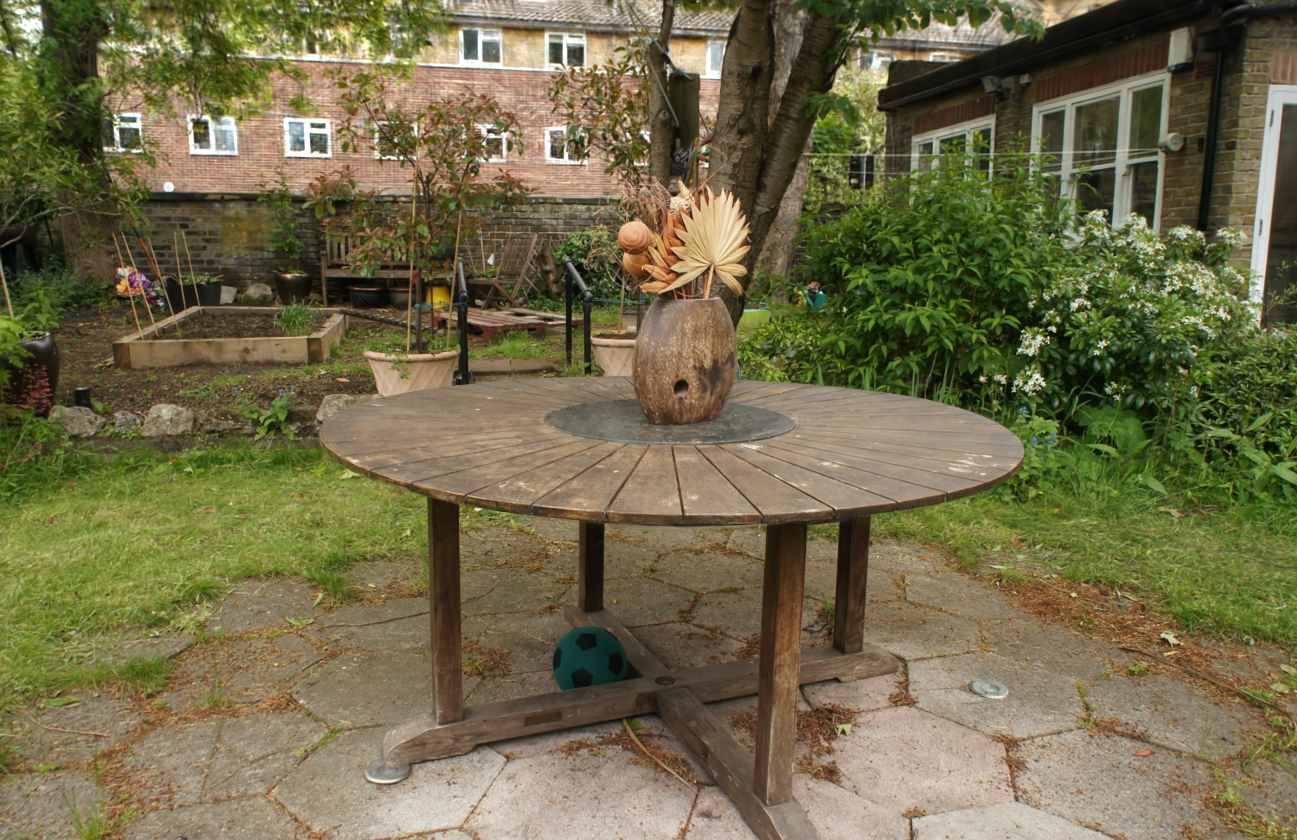}};
            \begin{scope}[x={(image.south east)},y={(image.north west)}]
                \draw[red,ultra thick,rounded corners] (0.01,0.70) rectangle (0.78,0.99);
            \end{scope}
        \end{tikzpicture}
        \label{fig:ours_high}
    }
    \caption{Qualitative visual comparison of our proposed method with MCMC approach~\cite{MCMC24} 
    on Garden scene with varying numbers of Gaussians. Our proposed method outperforms both visually
    and objectively MCMC thanks to a better reconstruction on background.}
    \label{fig:renderedImage}
\end{figure}
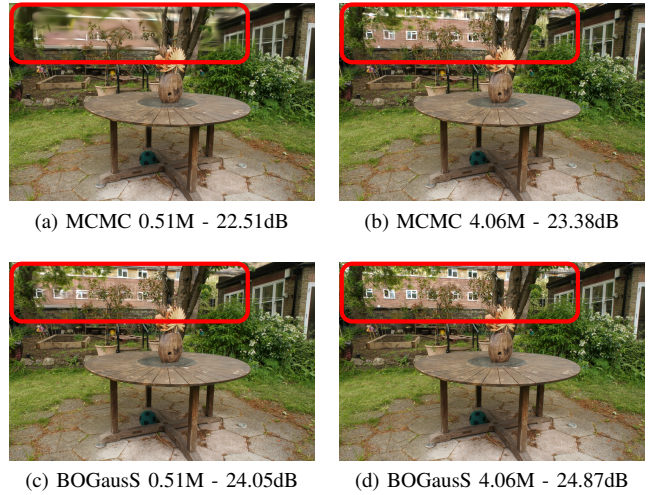

%% file: 2_related.tex
\section{Related Work}
\label{sec:relatedWork}

\subsection{3D Gaussian Splatting}

3D Gaussian Splatting \cite{kerbl3Dgaussians}  models the scene geometry as a set $\mathbb{G}$ of 3D Gaussians. 
Each Gaussian is characterized by its opacity $o_i \in [0,1]$, center's localization $P_i \in \mathbb{R}^3$ and a covariance
matrix in world space $\Sigma_i$ defining its spatial density 
$\mathcal{G}_i(X) = e^{-\frac{1}{2} (X-P_i)^T \Sigma_i^{-1}(X-P_i)}$. 
The covariance matrix $\Sigma_i$ is defined via sizes $\sigma_{i,a}, a \in [1..3]$ and a rotation.
To render an image $R^{(c)}$ associated to c-th camera, 
3DGS sorts the Gaussians in an approximate depth order based on the distance from their 
center $P_i$ to the image plane, and further applies alpha blending as initially proposed by Elliptical Weighted 
Averaging (EWA)~\cite{EWASplatting} for each pixel $p$:

\begin{IEEEeqnarray}{lrCl}
    &R^{(c)}(p) &=& \sum_{i \in \mathbb{G}} w^{(c)}_i \; c_i \nonumber \\
    \mathrm{with }&w_i &=& T^{(c)}_i \; o_i \; \mathcal{G}^{2D, c}_i(p) \nonumber  \\
    \mathrm{and } & T^{(c)}_i &=& \Pi_{j=1}^{i-1} (1 - o_j \; \mathcal{G}^{2D,c}_j(p))  
    \label{eq:rendering}
\end{IEEEeqnarray} 
where $c_i$ represents the view-dependent color modeled by spherical harmonics coefficients, and 
$\mathcal{G}^{2D,c}_j$ is the projected 2D Gaussian distribution of $\mathcal{G}_j$ through a local 
affine transformation. 

From~\eqref{eq:rendering} a loss function is defined based on a combination of distortion metrics such as 
$L_1$ or $L_2$ error, SSIM or other regularization metrics 
(e.g. depth regularization as in revised version of~\cite{kerbl3Dgaussians}, opacity and/or 
sizes decay~\cite{MCMC24}). This loss function is then minimized using 
Stochastic Gradient Descent techniques.

\subsection{Densification Optimization}

Finding the right number of Gaussians is important in order to get the best compromise 
between model complexity and quality rendering. 
Gaussians are generally initialized from sparse points (either issued from Structure-from-Motion pre-calibration 
or from random samples) and their density is iteratively refined by using pruning and splitting operations interleaved 
with loss minimization.
Several approaches have been proposed to optimize the selection of Gaussians to be pruned or split.

\textbf{Pruning.} 
In original 3DGS~\cite{kerbl3Dgaussians}, several criteria are used to prune Gaussians. 
Gaussians with low opacity or with too large size are pruned. \cite{MiniSplatting24,Eagles24} propose a 
 prioritization criterion based on the effective contribution of a Gaussian. 
 Effective contribution to  camera $c$ rendering is measured as 
$\mathcal{P}^{(c)}_i = \sum_p w^{(c)}_i(p)$. 

\textbf{Splitting.}
Original 3DGS~\cite{kerbl3Dgaussians} uses a 
prioritization mechanism based on visual positional gradient. 
Further studies have improved 3DGS performance by proposing various
prioritization criteria based on 
the marginal loss contribution \cite{RevDensif24},
on opacity \cite{MCMC24},
or on a mixture of metrics \cite{Taming24}.

Although providing significant improvements, these criteria remain empirically defined and
tuned (e.g. in \cite{Taming24} a dozen of hyperparameters have to be tuned).

%% file: 3_preliminary.tex
\section{Precision matrix derivation}
\label{sec:spatialPrecision}

One of the well-known artifacts of 3DGS is the 'needle' effect. Needle effect typically occurs 
when zooming in; thin Gaussians show up. \cite{MipSplatting24,FLOD24} propose solutions to 
limit these artifacts by considering either correction of low pass filtering as proposed in 
original EWA~\cite{EWASplatting}, or minimal size constraints for Gaussians. 
This highlights the need to control the minimal
size of Gaussians. \cite{FLOD24} suggests that this minimal size should be defined so that 
its projected size is not smaller than pixel size.
We propose here to rather use the precision accuracy on size estimation to define this minimal size.

Inspired from~\cite{Horaud16} using Variational Bayesian estimation in the context
of 3D reconstruction from multiple cameras, we model the reconstruction 
confidence on a Gaussian localization $P_i$. 
3D Gaussians' localizations are obtained from their projected 2D localization. 
So if 2D projected localizations
are subject to gaussian noise with variance ${\sigma^{(uv)}}^2$, 
then the 3D localization of the i-th Gaussian will also be 
subject to gaussian noise with covariance $\Sigma^{(XYZ)}_i$ given by:

\begin{equation}
    \label{eq:precision}
    {\Sigma^{(XYZ)}_i} ^{-1} = 
    \sum_c 
    \left(
        {z^{(c)}_i}^2 \; {\sigma^{(uv)}}^2 \; W^T_c \;
    \left(
        \begin{IEEEeqnarraybox*}[][c]{,c/c/c,}
        1 & 0 & 0 \\
        0 & 1 & 0 \\
        0 & 0 & 0        
        \end{IEEEeqnarraybox*}
    \right)
    \; W_c 
    \right)^{-1}
\end{equation}
where $W_c$ is the world transformation matrix associated to camera $c$ and $z^{(c)}_i$ is the distance of the Gaussian
along the z axis of the c-th camera. The same reasoning can be applied to derive the precision matrix for size,
which results in the same formula, $\Sigma^{(\sigma)}_i = \Sigma^{(XYZ)}_i$.

In this study, for simplicity, a scalar isotropic precision approximation is used 
to define confidence intervals for localization and sizes of the Gaussian as  
$\Delta_i = \alpha \; \text{tr}({\Sigma_i^{(XYZ)}}^{-1})^{-0.5}$ where $\alpha$ is a scaling factor set to 2. 
This scaling factor helps to accommodate with modeling approximations and also with the fact that in \eqref{eq:precision}
a Gaussian may not be seen by all cameras due to occlusions.

\section{Improved gradient descent}

\subsection{Proposed optimizer}

First we propose to modify the traditional Adam optimizer~\cite{Adam14} used as follows:
\begin{equation}
    \left\{
        \begin{matrix}
            \alpha_{s,t} & = & max(1/t, 1-\beta_s^{\star}) \\
            \widehat{g_0} & = & 0 \\
            \widehat{g_t} & = & \widehat{g_{t-1}} + \alpha_{1,t} (\nabla_{\theta} \mathcal{L}_t - \widehat{g_{t-1}})\\
            \widehat{g^2_t} & = & \widehat{g^2_{t-1}} + \alpha_{2,t} ((\nabla_{\theta} \mathcal{L}_t)^2 - \widehat{g^2_{t-1}})\\
            \theta_{t+1} & = & \theta_t + \eta_{\theta} \; \Delta_{\theta} \; \frac{\widehat{g_t}}{\widehat{g_t^2}+\epsilon}   
        \end{matrix}
    \right.
    \label{eq:adamNewState}
\end{equation}
where $\eta_{\theta}$ is the learning rate associated with  $\theta$  parameter to optimize. 
$(\beta^{\star}_s)$ are set as in traditional Adam optimizer. 
Modifications are related to the introduction of the $\Delta_{\theta}$ scaling factor in the update 
and the computation of filtered gradient states.
These modifications allow us to introduce positional-aware parameters' updates, a better handling of 
Gaussians not seen in some cameras and Gaussian states after splitting operations.

\subsection{Precision-aware parameters updates}

The confidence intervals derived in section ~\ref{sec:spatialPrecision} can also link the 3D variations
of localization and sizes to their impact on associated 2D projections and consequently to rendering variations. 
Thus, we propose to use the confidence interval $\Delta_i$ as $\Delta_{\theta}$ parameter in \eqref{eq:adamNewState} 
for localization and sizes parameters. For all other parameters we use $\Delta_{\theta}=1$. 

Consequently, sizes parameters $(\sigma_a)$ are linearly parameterized, unlike in 3DGS  implementations
where logarithmic parameterization is used (i.e. $\theta_{\sigma} = \log \sigma$). Logarithmic parameterization 
allows helping Gaussians sizes to evolve fast which is beneficial for far away Gaussians. 
Logarithmic parameterization of localization has also been proposed for 3D coordinates contraction 
(e.g.~\cite{SOG24}) for the same purpose.
Unfortunately these parameterizations are also more sensitive: any strong update may throw
away Gaussians or explode their sizes which often occurs in outdoor scenes. 
Our proposed precision-aware update is less prone to such artifacts and adapts to localization of 
the Gaussians and the camera setup. It is thus more general than other positional-aware mechanism such as proposed 
in \cite{PixelGS24} which implicitely assumes a specific camera setup.

\subsection{Sparse and unbiased states updates}

As in \cite{Taming24} we propose to use Sparse Adam to perform parameters update only on 
Gaussians that contribute to the rendered image being optimized. 
This allows to speed up code with a light performance decrease.
However, unlike biased estimator used in \cite{Taming24}, the BoGausS proposed modified Adam's optimizer 
defined by \eqref{eq:adamNewState} allows ensuring unbiased states 
$(\widehat{g_t}, \widehat{g^2_t})$ estimation
which prevents performance decrease and rather helps improve convergence 
(see ablation study in section \ref{sec:ablation}). 
Indeed instant $t$ in \eqref{eq:adamNewState} 
is Gaussian dependent and only incremented when a Gaussian is updated. $t$ then naturally refers 
 to Gaussian's lifespan. This additional state in optimizer is needed to ensure
unbiased states.

\subsection{States inheritance after Gaussian splitting}

When performing a splitting operation, the states of the Adam's optimizer are generally reset
to 0. 
In our approach, gradient states of Adam optimizer are partially propagated to Gaussian children via the 
following inheritance:
\begin{equation}
    \left\{
    \begin{matrix}
        t_{child} & = & \alpha_t \times t_{mother} \\
        \widehat{(g_{child})_t} & = & \alpha_g \times \widehat{(g_{mother})_t} \\
        \widehat{(g^2_{child})_t} & = & (\alpha_g)^2 \times \widehat{(g^2_{mother})_t} \\
    \end{matrix}
    \right.
\end{equation}
where $\alpha_t, \alpha_g$ are fading factors for inheritance of Gaussian lifespan
($t$ value used in~\eqref{eq:adamNewState}) and for Gaussian gradients states respectively. 
The lifespan fading factor $\alpha_t$ allows to control the lifespan inheritance and to control confidence in 
the reuse of the inherited filtered gradient. 
$\alpha_g$ allows to 
scale the gradient inheritance since the gradient amplitude naturally changes when considering 
a Gaussian child vs a Gaussian mother (e.g. since size of the splatted Gaussians will change, gradient
should be scaled). 
These fading parameters have been set empirically to $(\alpha_t=1, \alpha_g=0.2)$ for all datasets.

%% file: 4_methods.tex
\section{Improved Densification}
\label{sec:densif}

\subsection{Effective opacity-based Gaussian pruning}

For pruning operations, we propose to use the prioritization metric based on 
$\mathcal{P}^{(c)}_i = \frac {\sum_p w^{(c)}_i(p)}{S^{(c)}_i}$, 
where $S^{(c)}_i$ is the surface covered by the i-th Gaussian in c-th camera's image. 
$\mathcal{P}^{(c)}_i$ can be interpreted as the effective opacity of the i-th Gaussian.
The resulting prioritization metric for pruning is then defined as:
\begin{equation}
    \mathcal{P}_{prune}(i) = \max_c \mathcal{P}^{c}_i
\end{equation}
Gaussians to be pruned are the ones having $\mathcal{P}_{prune}$ value below a 
threshold $\tau_{prune}=0.02$. 
To prevent excessive pruning of Gaussians, 
we restrict the number of pruned Gaussians
to a fraction of the current total number of Gaussians (1\% in our experiments).
This pruning strategy naturally prunes Gaussians with too weak opacity and also those that 
may get out of view of training cameras. In BOGausS, it replaces opacity reset used in 3DGS.
Note that~\cite{Taming24} studied the use of a similar criterion but did not keep it because of  
inconsistent results for indoor and outdoor scenes.

\subsection{Density-preserving Gaussian splitting}

As mentioned in~\cite{MCMC24,RevDensif24} the original splitting operation causes 
perturbations in the loss function. 
Correction mechanisms on opacity and size parameters of split Gaussian children 
are then proposed to limit those perturbations. As perturbations also occur during cloning,
we choose to discard the cloning operations and rely only on splitting operations.

Inspired from~\cite{GaussianMixture03} we propose to modify the splitting operation
so that the resulting density after splitting is similar to the density of the Gaussian being split.
Splitting is performed along the longest principal axis, by shifting means by $\pm \alpha \sigma$, 
where $\alpha$ is a constant (set to 0.3 in our case) and $\sigma$ is the size along the 
splitting axis. To ensure that the resulting density of the sum of the two children is close
to the density of the mother Gaussian, we adjust the sizes and opacity of the split
Gaussians. This adaptation can be learned 
offline so that $(\sigma_{child}, o_{child}) = SplitAdaptation(\sigma_{mother}, o_{mother})$. 
\cite{densityControl24} uses a similar splitting mechanism, but does not take into account opacity variations.
This proposed splitting also prevents clusters of Gaussians as observed by~\cite{MiniSplatting24}.

\subsection{Distortion-based prioritization of Gaussians splitting}

We propose  
a prioritization criterion inspired from rate-distortion optimization~\cite{RDO98}. Gaussians to be split 
are the ones that will induce the highest gain in the PNSR quality metric. 

First we associate a reconstruction Squared Error $SE_{c,i}$ to each Gaussian that is defined as:
\begin{equation}
    SE_{c,i} = \sum_{p \in I_c} w_i^{(c)} \; ( R^{(c)}(p) - I^{(c)}(p))^2
\end{equation}
where $I^{(c)}$ corresponds to the ground-truth image acquired by the c-th camera.
As in~\cite{RevDensif24}, this formulation allows spreading error metric across Gaussians. 

When performing a splitting operation, the intent is to increase densification in the associated 
area in order to reduce reconstruction error. 
When splitting a Gaussian, the local reconstruction error will decrease from $SE_{c,i}$ to 
$\alpha_{split} SE_{c,i}$ with $0 < \alpha_{split} < 1$. 
The resulting variation in PSNR for that camera will then be:
\begin{equation}
    \begin{matrix}
        \Delta PSNR_{c,i} & = & 
        -10 ( \log_{10} SE_c^{(new)} - \log_{10} SE_c^{(prev)}) \\
        & = & -10 \log_{10} \frac { SE_c - (1-\alpha_{split})SE_{c,i}}{SE_c} \\
        & \simeq & \frac{10}{\ln(10)} \frac{(1-\alpha_{split})SE_{c,i}}{SE_c}
    \end{matrix}
\end{equation}
Considering that the Squared Error for camera $c$ can be expressed as $SE_c = \sum_j SE_{c,j}$
(assuming that the background contribution is negligible) and that $\alpha_{split}$ can be considered 
constant, then we can define the following 
prioritization densification criterion for Gaussian $i$:
\begin{equation}
    \mathcal{P}_{SNR}(i) = \sum_c \frac{SE_{c,i}}{\sum_j SE_{c,j}}
\end{equation}

Note that unlike the prioritization criterion proposed in \cite{RevDensif24} which is based 
only on the error $\mathcal{P}_{SE}(i)=\sum_c SE_{c,i}$, our proposed criterion is based on 
relative contributed error.

%% file: 5_results.tex
\section{Experiments}

\subsection{Experimental setup}

We run benchmarks on three established datasets: MipNerf360~\cite{barron2021mipnerf}, 
Tanks\&Temples~\cite{Tat17}, and Deep Blending~\cite{DeepBlending2018}, which contain 9, 2, 
and 2 scenes respectively. We use the same train/test split as the original 3DGS publication
and follow-up works. SSIM, PSNR and LPIPS metrics are evaluated. 
Training hyperparameters have same value for all scenes. 
We increase the interval between two densification operations to 500 iterations, 
allowing more time for Gaussians to stabilize 
and thereby get more reliable prioritization metrics, as proposed in~\cite{Taming24}.
Evaluations with and without exposure compensation are also proposed (based on technique proposed in~\cite{hierarchicalgaussians24}).

\subsection{Results}

\input{table_results}

\input{figure_SNR}

Comparative results are provided in Table~\ref{tab:results} and Fig.~\ref{fig:results}. 
Starting from the initial number of Gaussians obtained by vanilla 3DGS \cite{kerbl3Dgaussians} for each scene, 
we run our BOGausS solution for various targeted budgets that are 
fractions of 
initial number of Gaussians.
Results are compared with Taming~\cite{Taming24} and Mini-Splatting~\cite{MiniSplatting24} 
at various rates. 
Available code for MCMC~\cite{MCMC24} has been run  with aforementioned budget targets.
Looking at results in Table~\ref{tab:results} and Fig.~\ref{fig:results}, our method outperforms 
original 3DGS on all metrics. 
Even when using 5 times less Gaussians, our method exhibits higher quality. 
Compared to~\cite{MiniSplatting24,Taming24,MCMC24}, 
BOGausS consistently provides better PSNR, and similar SSIM and LPIPS values.
Compared to MCMC, BOGausS may provide only slightly higher PSNR,  
but it provides significantly higher subjective quality as shown in Fig.~\ref{fig:renderedImage}.
Typically, background content is richer with finer details thanks
to denser Gaussians in these areas without sacrificing reconstruction of foreground content. 
The discrepancy between subjective quality gains and objective metrics may be due to camera calibrations issues.
For instance when applying exposure correction (such as proposed in~\cite{hierarchicalgaussians24}), large gains in 
objective metrics can be observed (see bottom part of Table~\ref{tab:results}). 
Pose estimation is also a problem: some misalignment can be observed between 
rendered images and ground-truth images.

\subsection{Ablation study}
\label{sec:ablation}
\begin{table}[ht]
    \caption{Ablation of BOGausS's main components. Tested on Garden scene with 1.01M Gaussians.}
    \label{tab:ablations}
    \centering
    \begin{tabular}{l|ccc}
        & SSIM $\uparrow$ & PSNR $\uparrow$ & LPIPS $\downarrow$ \\
        \hline
        BOGausS &  \cellcolor{Apricot} 0.861 & \cellcolor{Salmon} 27.81 & \cellcolor{Apricot} 0.132 \\
        \hline
        no Sparse Adam & 0.854 & 27.49 & 0.140 \\      
        no states inheritance at split & 0.859 & \cellcolor{Apricot} 27.77 & 0.134 \\
        no scaled updates & 0.854 & 27.51 & 0.140 \\
        no effective opacity pruning & 0.855 & \cellcolor{Apricot} 27.77 &  0.1411\\
        no SNR split prioritization & \cellcolor{Salmon} 0.863 & 27.66 & \cellcolor{Salmon} 0.126 \\
        \hline       
    \end{tabular}
    
\end{table}

Table~\ref{tab:ablations} examines the effect of individually removing several of our contributions. 
The analysis is performed on Garden sequence with an intermediate number of Gaussians. 
Removing any of these contributions degrades PSNR performance. Removing inherited optimizer
states during splitting operations does not bring much degradation, but we have observed that
it introduces more variations during loss minimization. It also comes with more Gaussians being
pruned due to more instabilities. Replacing SNR prioritization with MCMC prioritization mechanism for densification 
provides slightly better SSIM and LPIPS values, but lower PSNR. 
Moreover, as can be observed in Fig.~\ref{fig:renderedImage}, it does not
have a good behavior in background areas with significant visual artifacts.

%% file: table_results.tex
\begin{table*}[!t]
    \caption{Quantitative comparison of other methods with our technique. 
    Upper set of methods in each half targets same number of Gaussians than vanilla 3DGS, lower set targets lower number of Gaussians.
    \colorbox{Salmon}{Best} and \colorbox{Apricot}{Second Best} results are highlighted for each dataset and category}
    \label{tab:results}
    \centering
    \begin{tabular}{l|cccc|cccc|cccc}
        & \multicolumn{4}{c|}{Mip-Nerf 360} & \multicolumn{4}{c|}{Tanks\&Temples} & \multicolumn{4}{c}{Deep Blending} \\
        & SSIM $\uparrow$ & PSNR $\uparrow$ & LPIPS $\downarrow$ & \#G. & SSIM $\uparrow$ & PSNR $\uparrow$ & LPIPS $\downarrow$ & \#G. & SSIM $\uparrow$ & PSNR $\uparrow$ & LPIPS $\downarrow$ & \#G. \\
        \hline
        \multicolumn{13}{c}{Without Exposure Correction} \\ 
        \hline
        3DGS \cite{kerbl3Dgaussians} & \input{3dgs-1.0.tex} \\
        Mini-SplattingD \cite{MiniSplatting24} & \input{minisplatting-D.tex} \\
        Taming (big)\cite{Taming24} & \input{taming-1.00.tex} \\ 
        MCMC \cite{MCMC24} & \input{mcmc-1.00.tex} \\
        Ours (Big) & \input{bogauss-30k-fast-1.00.tex} \\
        \hline 
        Taming (small) \cite{Taming24} & \input{taming-0.10.tex} \\
        Mini-Splatting \cite{MiniSplatting24} & \input{minisplatting.tex} \\
        Ours (Light) & \input{bogauss-30k-fast-0.10.tex} \\
        Ours (Medium) & \input{bogauss-30k-fast-0.40.tex} \\
        \hline
        \multicolumn{13}{c}{With Exposure Correction} \\ 
        \hline
        3DGS \cite{kerbl3Dgaussians} & \input{3dgs-exp.tex} \\
        MCMC \cite{MCMC24} & \input{mcmc-exp-30k-1.00.tex} \\
        Ours (Big) & \input{bogauss-exp-30k-fast-1.00.tex} \\
        \hline 
        Ours (Light) & \input{bogauss-exp-30k-fast-0.10.tex} \\
        Ours (Medium) & \input{bogauss-exp-30k-fast-0.40.tex} \\
        \hline
        
    \end{tabular}

\end{table*}

%% file: 3dgs-1.0.tex
0.816 & 27.57 & 0.215 & 3.31 & 0.853 & 23.79 & 0.169 & 1.84 & 0.906 & 29.69 & 0.238 & 2.81

%% file: minisplatting-D.tex
0.831 & 27.51 & 0.176 & 4.69 & 0.853 & 23.23 & \cellcolor{Salmon} 0.140 & 4.28 & 0.906 & 29.88 & \cellcolor{Salmon} 0.211 & 4.63

%% file: taming-1.00.tex
0.822 & 27.79 & 0.205 & 3.31 & 0.851 & 24.04 & 0.170 & 1.84 & \cellcolor{Apricot} 0.907 & \cellcolor{Apricot} 30.14 & 0.235 & 2.81

%% file: mcmc-1.00.tex
\cellcolor{Salmon} 0.846 & \cellcolor{Apricot} 28.30 & \cellcolor{Apricot} 0.174 & 3.21 & \cellcolor{Salmon} 0.871 & \cellcolor{Apricot} 24.58 & \cellcolor{Apricot} 0.147 & 1.83 & 0.905 & 29.34 & 0.238 & 2.80

%% file: bogauss-30k-fast-1.00.tex
\cellcolor{Apricot} 0.844 & \cellcolor{Salmon} 28.40 & \cellcolor{Salmon} 0.172 & 3.21 & \cellcolor{Apricot} 0.866 & \cellcolor{Salmon} 24.78 & 0.176 & 1.83 & \cellcolor{Salmon} 0.915 & \cellcolor{Salmon} 30.41 & \cellcolor{Apricot} 0.230 & 2.80

%% file: taming-0.10.tex
0.799 & 27.29 & 0.253 & 0.63 & 0.835 & 23.89 & 0.207 & 0.29 & 0.902 & 29.89 & 0.263 & 0.27

%% file: minisplatting.tex
0.822 & 27.34 & 0.217 & 0.49 & 0.835 & 23.18 & 0.202 & 0.20 & 0.908 & 29.98 & 0.253 & 0.35

%% file: bogauss-30k-fast-0.10.tex
0.805 & 27.21 & 0.251 & 0.32 & 0.830 & 23.72 & 0.250 & 0.18 & 0.908 & 30.18 & 0.259 & 0.28

%% file: bogauss-30k-fast-0.40.tex
0.836 & 28.21 & 0.192 & 1.28 & 0.857 & 24.62 & 0.199 & 0.73 & 0.913 & 30.37 & 0.237 & 1.12

%% file: 3dgs-exp.tex
0.818 & 27.74 & 0.214 & 3.31 & 0.862 & 24.94 & \cellcolor{Salmon} 0.164 & 1.84 & 0.912 & 30.59 & \cellcolor{Apricot} 0.236 & 2.81

%% file: mcmc-exp-30k-1.00.tex
\cellcolor{Apricot} 0.840 & 28.36 & \cellcolor{Apricot} 0.182 & 3.21 & \cellcolor{Salmon} 0.877 & \cellcolor{Salmon} 26.00 & \cellcolor{Apricot} 0.170 & 1.83 & 0.908 & 30.60 & 0.237 & 2.80

%% file: bogauss-exp-30k-fast-1.00.tex
\cellcolor{Salmon} 0.844 & \cellcolor{Salmon} 28.82 & \cellcolor{Salmon} 0.171 & 3.21 & \cellcolor{Apricot} 0.872 & \cellcolor{Apricot} 25.99 & \cellcolor{Apricot} 0.170 & 1.83 & \cellcolor{Salmon} 0.915 & \cellcolor{Salmon} 31.25 & \cellcolor{Salmon} 0.229 & 2.80

%% file: bogauss-exp-30k-fast-0.10.tex
0.804 & 27.54 & 0.251 & 0.32 & 0.837 & 24.85 & 0.244 & 0.18 & 0.908 & 30.96 & 0.259 & 0.28

%% file: bogauss-exp-30k-fast-0.40.tex
0.836 & \cellcolor{Apricot} 28.59 & 0.191 & 1.28 & 0.862 & 25.79 & 0.193 & 0.73 & \cellcolor{Apricot} 0.914 & \cellcolor{Apricot} 31.24 & 0.237 & 1.12

%% file: figure_SNR.tex
\begin{figure*}[t]
    \centering

    \begin{tikzpicture}

        \begin{axis}[
                name=ax1,
                title=Mip-Nerf 360,
                grid= both,
                width=0.33*\linewidth,
                height=5cm,
                xlabel = {Number of Gaussians (Million)},
                ylabel = {PSNR (dB)} ,
                xmin = 0.0, xmax = 4,
                xlabel near ticks, minor x tick num=1,
                ymin = 26, ymax = 29,
                ylabel near ticks, minor y tick num=3, ytick distance={1},
                title style={yshift=-0.75ex},
                ylabel shift=-0.15cm,
                legend style={font=\small, at={(-0.1,0.-0.2)},anchor=north west,legend columns=2},
                ]

            \addplot[thick, bluecharted, only marks, mark=*, mark size=2pt] table[x index=0, y index=1] {mip360-3dgs.csv};
            \addplot[thick, greenarm, only marks, mark=o, mark size=2pt] table[x index=0, y index=1] {mip360-3dgs-exp.csv};
            \addplot[thick, pinkcharted, only marks, mark=triangle*, mark size=2pt] table[x index=0, y index=1] {mip360-minisplatting.csv};
            \addplot[thick, dashed, purplecharted, mark=pentagon*, mark size=2pt, mark options={solid}] table[x index=0, y index=1] {mip360-taming.csv};
            \addplot[thick, dashed, greencharted, mark=diamond*, mark size=2pt, mark options={solid}] table[x index=0, y index=1] {mip360-mcmc.csv};
            \addplot[thick, dashed, greencharted, mark=diamond, mark size=2pt, mark options={solid}] table[x index=0, y index=1] {mip360-mcmc-exp-30k.csv};
            \addplot[thick, dashed, orangecharted, mark=square, mark size=2pt, mark options={solid}] table[x index=0, y index=1] {mip360-bogauss-exp-30k-fast.csv};
            \addplot[thick, dashed, orangecharted, mark=square*, mark size=2pt, mark options={solid}] table[x index=0, y index=1] {mip360-bogauss-30k-fast.csv};
            \draw[dashed, thick, black] (axis cs:0., 27.57) -- (axis cs:10, 27.57);

        \end{axis}

        \begin{axis}[
            at={(ax1.south east)},
            name=ax2,
            xshift=1.5cm,
            title=Tanks\&Temples,
            grid= both,
            width=0.33*\linewidth,
            height=5cm,
            xlabel = {Number of Gaussians (Million)},
            ylabel = {PSNR (dB)} ,
            xmin = 0, xmax = 2,
            xlabel near ticks, minor x tick num=1,
            ymin = 23, ymax = 26.5, ylabel near ticks, minor y tick num=3, ytick distance={1},
            title style={yshift=-0.75ex},
            ylabel shift=-0.15cm,
            legend style={font=\small, at={(0,-0.1)},anchor=north east,legend columns=3},
            ]

            \addplot[thick, bluecharted, only marks, mark=*, mark size=2pt] table[x index=0, y index=1] {tat-3dgs.csv}; 
            \addplot[thick, bluecharted, only marks, mark=o, mark size=2pt] table[x index=0, y index=1] {tat-3dgs-exp.csv};
            \addplot[thick, pinkcharted, only marks, mark=triangle*, mark size=2pt] table[x index=0, y index=1] {tat-minisplatting.csv};
            \addplot[thick, dashed, purplecharted, mark=pentagon*, mark size=2pt, mark options={solid}] table[x index=0, y index=1] {tat-taming.csv};
            \addplot[thick, dashed, greencharted, mark=diamond*, mark size=2pt, mark options={solid}] table[x index=0, y index=1] {tat-mcmc.csv};
            \addplot[thick, dashed, greencharted, mark=diamond, mark size=2pt, mark options={solid}] table[x index=0, y index=1] {tat-mcmc-exp-30k.csv};
            \addplot[thick, dashed, orangecharted, mark=square, mark size=2pt, mark options={solid}] table[x index=0, y index=1] {tat-bogauss-exp-30k-fast.csv};
            \addplot[thick, dashed, orangecharted, mark=square*, mark size=2pt, mark options={solid}] table[x index=0, y index=1] {tat-bogauss-30k-fast.csv};
            \draw[dashed, thick, black] (axis cs:0., 23.79) -- (axis cs:10, 23.79);
        \end{axis}

        \begin{axis}[
            at={(ax2.south east)},
            name=ax3,
            xshift=1.5cm,
            title=Deep Blending,
            grid= both,
            width=0.33*\linewidth,
            height=5cm,
            xlabel = {Number of Gaussians (Million)},
            ylabel = {PSNR (dB)} ,
            xmin = 0, xmax = 3,
            xlabel near ticks, minor x tick num=1,
            ymin = 28, ymax = 31.5,
            ylabel near ticks, minor y tick num=3, ytick distance={1},
            title style={yshift=-0.75ex},
            ylabel shift=-0.15cm,
            legend style={font=\small, at={(1.,-0.3)},anchor=north east,legend columns=5},
            legend cell align={left}, 
            ]

            \addplot[thick, bluecharted, only marks, mark=*, mark size=2pt] table[x index=0, y index=1] {db-3dgs.csv};
            \addlegendentry{\sf \small 3DGS}
            \addplot[thick, pinkcharted, only marks, mark=triangle*, mark size=2pt] table[x index=0, y index=1] {db-minisplatting.csv};
            \addlegendentry{\sf \small  Mini-Splatting } 
            \addplot[thick, dashed, purplecharted, mark=pentagon*, mark size=2pt, mark options={solid}] table[x index=0, y index=1] {db-taming.csv};
            \addlegendentry{\sf \small Taming}
            \addplot[thick, dashed, greencharted, mark=diamond*, mark size=2pt, mark options={solid}] table[x index=0, y index=1] {db-mcmc.csv};
            \addlegendentry{\sf \small MCMC}
            \addplot[thick, dashed, orangecharted, mark=square*, mark size=2pt, mark options={solid}] table[x index=0, y index=1] {db-bogauss-30k-fast.csv};
            \addlegendentry{\sf \small BOGausS (Ours)}
            \addplot[thick, bluecharted, only marks, mark=o, mark size=2pt] table[x index=0, y index=1] {db-3dgs-exp.csv};
            \addlegendentry{\sf \small 3DGS - Exp. comp.}
            \addplot[thick, dashed, greencharted, mark=diamond, mark size=2pt, mark options={solid}] table[x index=0, y index=1] {db-mcmc-exp-30k.csv};
            \addlegendentry{\sf \small MCMC - Exp. comp.}
            \addplot[thick, dashed, orangecharted, mark=square, mark size=2pt, mark options={solid}] table[x index=0, y index=1] {db-bogauss-exp-30k-fast.csv};
            \addlegendentry{\sf \small BOGausS (Ours) - Exp. comp.}
            \draw[dashed, thick, black] (axis cs:0., 29.69) -- (axis cs:10, 29.69);
            \node [black, above, xshift=0.4cm] at (axis cs:5, 27.57){\sf \small 3DGS};
        \end{axis}

    \end{tikzpicture}

    \caption{Comparative PSNR performances vs number of Gaussians. 
    Unfilled-marks plots use exposure compensation while filled-marks plots don't.}
    \label{fig:results}
\end{figure*}
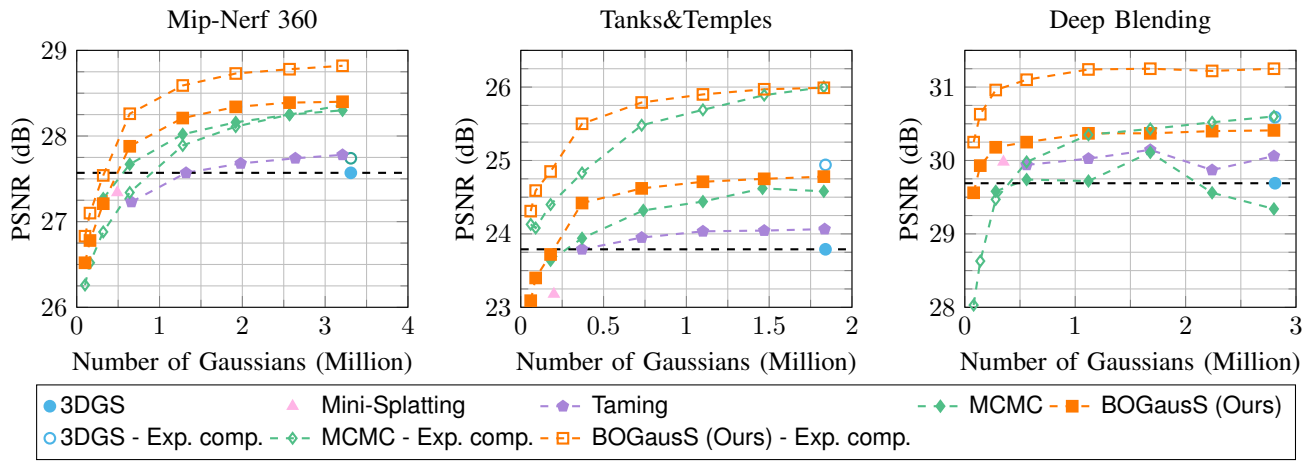

%% file: 6_conclusion.tex
\section{Conclusion}

We have presented a Better Optimized Gaussian Splatting (BOGausS) algorithm for generating high-quality novel view 
synthesis. 
Our approach provides significant improvements over previous work allowing to generate Gaussian splatting
content with fewer primitives and higher quality. Compared to original 3DGS, BOGausS provides similar quality
with 5 times less Gaussians, even 10 times less Gaussians if using exposure correction.
Allowing graceful degradation with reduced number of Gaussians is especially
beneficial for low-resources systems.

First observations show that some additional gain could be obtained by tuning the new introduced hyperparameters. 
We let these optimizations for further studies.